\documentclass[conference]{IEEEtran}
\IEEEoverridecommandlockouts
\usepackage{cite}
\usepackage{amsmath,amssymb,amsfonts}
\usepackage{graphicx}
\usepackage{textcomp}
\usepackage{xcolor}
\usepackage{booktabs}
\usepackage{url}
\usepackage{hyperref}
\hypersetup{hidelinks}
\usepackage{tikz}
\usetikzlibrary{arrows.meta,positioning,fit,backgrounds,calc,shapes.geometric}

\begin{document}

\title{Component Type, Not Reconstruction Error, Predicts Attention Quantization Sensitivity}

\author{
\IEEEauthorblockN{
Kasun Dewage,
Marianna Pensky,
and Suranadi De Silva
}

\IEEEauthorblockA{
\textit{University of Central Florida}\\
Orlando, Florida, USA\\
\texttt{\{KasunTharuka.Dewage, Marianna.Pensky, su966204\}@ucf.edu}
}
\thanks{Code is publicly available at \url{https://github.com/Kasun-Dewage/What-predicts-Quantization-Sensitivity-2026}.}
\thanks{Accepted as a regular paper at IEEE ICMLA 2026. \textcopyright~2026 IEEE. Personal use of this material is permitted. Permission from IEEE must be obtained for all other uses, in any current or future media, including reprinting/republishing this material for advertising or promotional purposes, creating new collective works, for resale or redistribution to servers or lists, or reuse of any copyrighted component of this work in other works.}
}

\maketitle

%
%
\definecolor{stageblue}{RGB}{31,78,121}
\definecolor{stagebg}{RGB}{222,235,247}
\definecolor{vcol}{RGB}{197,90,17}
\definecolor{loopred}{RGB}{192,0,0}
\definecolor{metricbg}{RGB}{237,237,237}

\begin{figure*}[!t]
\centering
\begin{tikzpicture}[
  font=\small,
  >={Stealth[length=2.2mm]},
  stage/.style={draw=stageblue,line width=0.6pt,fill=stagebg,rounded corners=2pt,
                align=center,inner sep=4pt,minimum height=9mm},
  proj/.style={draw,rounded corners=1pt,minimum width=6.5mm,minimum height=5mm,
               inner sep=1pt,font=\footnotesize\bfseries},
  metric/.style={draw=black!55,fill=metricbg,rounded corners=1pt,align=left,
                 inner sep=3pt,font=\footnotesize,minimum height=6mm},
  outlab/.style={draw=stageblue,line width=0.6pt,fill=white,rounded corners=2pt,
                 align=center,inner sep=3pt,font=\footnotesize},
  arr/.style={->,line width=0.7pt},
]

\node[stage,text width=20mm] (model)
     {\textbf{9 open-weight}\\\textbf{LLMs}\\[1pt]
      \footnotesize OPT, GPT-J,\\\footnotesize LLaMA-1/2/3,\\\footnotesize Mistral, Qwen2.5};
\node[stage,below=4mm of model,text width=20mm] (base)
     {\textbf{Baseline PPL}\\[1pt]\footnotesize WikiText-2\\
      \footnotesize 16{,}384 tok,\\\footnotesize 1{,}024-tok blocks};
\node[stage,above=4mm of model,text width=20mm] (calib)
     {\textbf{Calib.\ stats}\\[1pt]\footnotesize fwd hooks\\\footnotesize $\to\ \widehat{\mathbb{E}}[x_j^2]$};

\node[stage,right=14mm of model,text width=24mm] (attn)
      {\textbf{One projection}\\\textbf{at a time}\\[3pt]
       \footnotesize Q, K, V, O\\\footnotesize $\times$ every layer};
\node[proj,fill=black!12,draw=black!55,below=7mm of attn.south,xshift=-12mm] (Q) {Q};
\node[proj,fill=black!12,draw=black!55,right=1.4mm of Q] (K) {K};
\node[proj,fill=vcol!28,draw=vcol,line width=1pt,right=1.4mm of K] (V) {V};
\node[proj,fill=black!12,draw=black!55,right=1.4mm of V] (O) {O};
\node[font=\scriptsize\itshape,below=1mm of K.south,xshift=3mm] {selected weight $W$};

\node[stage,right=15mm of attn,text width=28mm] (quant)
     {\textbf{Quantize} $W\!\to\!W_q$\\[1pt]
      \footnotesize RTN \emph{or} GPTQ\\
      \footnotesize $b\in\{4,3\}$ bit, gs$=128$};
\node[stage,below=4mm of quant,text width=28mm] (evalppl)
     {\textbf{Re-evaluate PPL}\\[1pt]\footnotesize all others full precision\\
      $\Delta\text{PPL}=\text{PPL}_q-\text{PPL}_{\text{base}}$};
\node[outlab,below=4mm of evalppl,text width=28mm,fill=loopred!8,draw=loopred]
     (restore) {\textcolor{loopred}{\textbf{Restore}} original $W$;\\ advance to next projection};

\node[metric,above=4mm of quant,text width=30mm] (metrics)
     {\textbf{Record 3 signals}\\[2pt]
      recon: $\|W\!-\!W_q\|_F/\|W\|_F$\\[1pt]
      act.-wtd: $\hat S=\frac1m\sum_{ij}(W_{ij}\!-\!W_{q,ij})^2\widehat{\mathbb{E}}[x_j^2]$\\[1pt]
      functional: $\Delta\text{PPL}$};

\node[stage,right=15mm of quant,text width=24mm] (dataset)
      {\textbf{3{,}808}\\\textbf{measurements}\\[2pt]
       \footnotesize 2{,}144 RTN (9 mdl)\\
       \footnotesize 1{,}664 GPTQ (7 mdl)};
\node[outlab,below=4mm of dataset,text width=24mm] (a1)
     {recon.\ error is weak:\\ median $R^2=0.044$};
\node[outlab,below=4mm of a1,text width=24mm] (a2)
     {\textcolor{vcol}{\textbf{V dominates 7/9}}\\ (38--51\% of total positive $\Delta$PPL)};
\node[outlab,above=4mm of dataset,text width=24mm] (a3)
     {$\hat S$ beats recon on V:\\ $R^2$ 0.20 vs.\ 0.06};

\draw[arr] (model) -- (attn);
\draw[arr,dashed] (calib.east) to[out=0,in=150] ($(attn.north west)+(2mm,-1mm)$);
\draw[arr,dashed] (base.east)  to[out=0,in=210] ($(attn.south west)+(2mm,1mm)$);
\draw[arr] (attn) -- (quant);
\draw[arr] (quant) -- (evalppl);
\draw[arr] (quant.north) -- (metrics.south);
\draw[arr] (quant) -- (dataset);

\coordinate (loopY) at ($(restore.south)+(0,-5mm)$);
\draw[arr,loopred,dashed,line width=0.8pt]
   (restore.south) |- (loopY) -| (attn.south);
\node[loopred,font=\scriptsize\itshape] at ($(loopY)+(0,2.2mm)$)
   {repeat for every projection $\times$ layer $\times$ bit-width};

\end{tikzpicture}
\caption{Overview of the experimental pipeline. For each of nine open-weight
models we compute a baseline WikiText-2 perplexity and capture per-channel
activation second moments, then quantize \emph{one attention projection at a
time} (Q, K, V, or O at every layer) to $b\in\{4,3\}$ bits under RTN or GPTQ
while all other weights stay at full precision. Each trial records the relative
reconstruction error, the activation-weighted quantization error
$\hat S$, and the perplexity change $\Delta$PPL, after which the original weight
is restored. The sweep yields 3{,}808 measurements (2{,}144 RTN across nine
models; 1{,}664 GPTQ across seven). Key results: reconstruction error is a weak
within-component predictor of $\Delta$PPL (median $R^2=0.044$); V projections
dominate sensitivity in 7 of 9 models (38--51\% of total positive $\Delta$PPL); and $\hat S$
predicts V-projection sensitivity far better than reconstruction error
(median $R^2$ 0.20 vs.\ 0.06).}
\label{fig:overview}
\end{figure*}
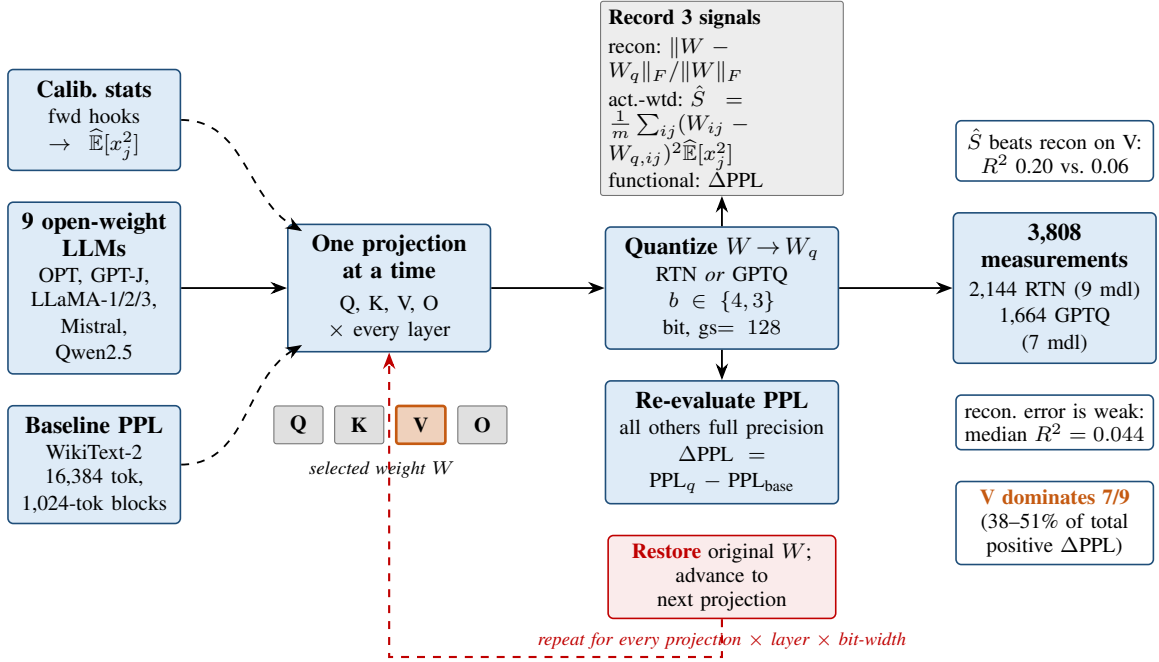

\begin{abstract}
Many post-training quantization (PTQ) methods use layer-wise reconstruction, second-order proxy objectives, or activation-aware transformations to reduce quantization-induced error. Whether that error signal predicts the downstream functional impact of quantizing an individual attention projection has not been directly characterized. We sweep nine open-weight language models (1.3B--8B parameters; OPT, GPT-J, LLaMA-1/2/3, Mistral, Qwen 2.5) and quantize one attention projection at a time under round-to-nearest (RTN) and, for seven models, GPTQ at 3 and 4 bits, recording reconstruction error, perplexity change, and per-projection activation-weighted quantization error for 3,808 distinct measurements. We find: (1)~within a given component type (Q, K, V, or O), reconstruction error explains less than 10\% of the variance in perplexity sensitivity in 27 of 36 cases under RTN, with median $R^2 = 0.044$; (2)~both component type and layer identity explain more variance than reconstruction error in all 9 models, with layer identity the strongest predictor in 7 of 9 models and component type strongest in the remaining 2; (3)~value (V) projections are the most commonly dominant component, accounting for 38--51\% of total positive $\Delta\text{PPL}$ in seven of nine models; (4)~the dominant component is broadly preserved between RTN and GPTQ (5 of 7 cases); and (5)~activation-weighted quantization error is a moderately better within-component predictor than reconstruction error for V projections specifically (median $R^2$ of 0.20 vs.\ 0.06). These findings indicate that relative weight reconstruction error alone is insufficient for sensitivity-aware bit allocation, and that V projections merit dedicated consideration in mixed-precision schemes.
\end{abstract}


\section{Introduction}

Modern post-training quantization (PTQ) methods for large language models use calibration-based objectives or transformations to reduce quantization-induced error. GPTQ~\cite{frantar2023gptq}, QuIP~\cite{chee2023quip}, and QuIP\#~\cite{tseng2024quipsharp} directly optimize second-order or quadratic proxy objectives, while AWQ~\cite{lin2024awq}, SqueezeLLM~\cite{kim2023squeezellm}, OmniQuant~\cite{shao2024omniquant}, and SmoothQuant~\cite{xiao2023smoothquant} use activation-aware scaling, dense-and-sparse decomposition, calibration, or equivalent transformations to preserve model behavior under low-bit quantization. A common proxy objective for analyzing such methods is the layer-wise reconstruction objective:
\begin{equation}
\min_{W_q} \|WX - W_q X\|_F^2,
\label{eq:recon}
\end{equation}
where $W$ is a full-precision weight matrix, $W_q$ its quantized counterpart, and $X$ a batch of calibration activations. The implicit assumption is that minimizing this reconstruction error at each layer independently approximately minimizes the downstream effect on model quality. APTQ~\cite{guan2024aptq} and QAQ~\cite{dong2024qaq} have observed that attention projections vary in sensitivity and have proposed Hessian-trace-based mixed-precision allocation; these works use richer signals than reconstruction error alone. Recent work on quantization error propagation~\cite{arai2025qep} has further highlighted the limitations of independent layer-wise optimization.

In this paper we directly characterize the predictive value of the reconstruction-error signal at the level of individual attention projections. For each of nine open-weight models, we quantize \emph{one attention projection at a time} (Q, K, V, or O at every layer) under RTN and, for seven models, GPTQ, and we measure (i)~the relative reconstruction error $\|W - W_q\|_F / \|W\|_F$, (ii)~the resulting perplexity change $\Delta\text{PPL}$ on WikiText-2 with all other weights at full precision, and (iii)~the per-projection activation-weighted quantization error. This produces 3,808 distinct measurements (2,144 RTN measurements across nine models and 1,664 GPTQ measurements across seven models) and lets us ask: does reconstruction error predict $\Delta\text{PPL}$? Does component type? Does layer identity? Figure~\ref{fig:overview} provides a visual overview.

The short answers are: reconstruction error is a weak within-component predictor in 75\% of cases under RTN (median $R^2=0.044$), both component type and layer identity explain more variance than reconstruction error in all 9 models, and layer identity is the strongest individual predictor in 7 of 9 models with component type strongest in the remaining 2. Among components, V projections are dominant in seven of the nine models, and the dominance pattern is largely preserved between RTN and GPTQ. We close by noting that activation-weighted quantization error is a noticeably better predictor than reconstruction error for V projections (median $R^2 = 0.20$ across V cells under RTN, vs.\ 0.06 for reconstruction error).

\textbf{Contributions.} (1)~A 3,808-measurement empirical map of single-projection quantization sensitivity across nine RTN models and seven GPTQ models. (2)~Direct evidence that relative weight reconstruction error, a commonly reported PTQ diagnostic, is a weak within-component predictor of perplexity sensitivity. (3)~A V-projection-centric dominance picture that holds across architecture families and across PTQ methods, contrary to the K-centric reading suggested by some prior work. (4)~Identification of activation-weighted quantization error as a substantially stronger V-projection predictor than reconstruction error, motivating its broader use.

\section{Related Work}

\textbf{Layer-wise quantization sensitivity.} HAWQ~\cite{dong2019hawq} and HAWQ-V2~\cite{dong2020hawqv2} use Hessian information to determine per-block bit-widths. They operate at the transformer-block level; we operate at the individual-projection level within attention.

\textbf{Attention-aware mixed precision.} APTQ~\cite{guan2024aptq} proposes mixed-precision allocation within attention blocks using the Hessian trace, and reports that K projections in LLaMA are particularly sensitive. QAQ~\cite{dong2024qaq} proves K and V caches have distinct quantization sensitivities and proposes separate strategies for each; its focus is the KV cache rather than projection weights. SensiBoost/KurtBoost~\cite{zhang2025sensiboost} use activation sensitivity and weight kurtosis. Spike-aware methods~\cite{maisonnave2025spike} target FFN down-projections with large activation spikes. ResQ~\cite{saxena2024resq} introduces mixed-precision quantization with low-rank residuals for improved accuracy. Our contribution is complementary: we characterize how informative the reconstruction-error signal itself is, across a wider span of architectures, and ask whether component-level effects survive a switch from RTN to GPTQ.

\textbf{Post-training quantization methods.} GPTQ~\cite{frantar2023gptq} applies optimal-brain quantization row-by-row; AWQ~\cite{lin2024awq} protects salient channels by activation magnitude; QuIP~\cite{chee2023quip} uses incoherence processing; QuIP\#~\cite{tseng2024quipsharp} extends this with Hadamard transforms and lattice codebooks; SmoothQuant~\cite{xiao2023smoothquant} migrates quantization difficulty from activations to weights; OmniQuant~\cite{shao2024omniquant} learns omnidirectional quantization parameters. These methods differ in whether they use explicit reconstruction losses, second-order proxy objectives, activation-aware scaling, or equivalent transformations. Our question is narrower: whether relative weight reconstruction error, when measured at the attention-projection level, correlates with $\Delta$PPL.

\textbf{Outlier-aware quantization.} LLM.int8()~\cite{dettmers2022gptint8}, SpQR~\cite{dettmers2023spqr}, and Super Weight~\cite{yu2024superweight} address \emph{which weights} are critical without explicitly distinguishing attention component types.

\section{Experimental Setup}
\label{sec:setup}

\subsection{Models}

We evaluate nine open-weight models (Table~\ref{tab:models}). The set spans multi-head attention (MHA), grouped-query attention (GQA), and five model families (OPT, GPT-J, LLaMA, Mistral, Qwen) ranging from 1.3B to 8B parameters.

\begin{table}[t]
\centering
\caption{Models evaluated.}
\label{tab:models}
\small
\begin{tabular}{llrl}
\toprule
\textbf{Model} & \textbf{Family} & \textbf{Params} & \textbf{Attn.} \\
\midrule
OPT-1.3B & OPT & 1.3B & MHA \\
OPT-6.7B & OPT & 6.7B & MHA \\
GPT-J-6B & GPT-J & 6B & MHA \\
LLaMA-1-7B & LLaMA & 7B & MHA \\
LLaMA-2-7B & LLaMA & 7B & MHA \\
LLaMA-3-8B & LLaMA & 8B & GQA \\
Mistral-7B & Mistral & 7B & GQA \\
Qwen2.5-1.5B & Qwen & 1.5B & GQA \\
Qwen2.5-7B & Qwen & 7B & GQA \\
\bottomrule
\end{tabular}
\end{table}

\subsection{Quantization Methods}

We use \emph{symmetric per-row} RTN with group size 128, and GPTQ~\cite{frantar2023gptq} with the same group size and 2,048 calibration tokens drawn from C4. For each weight matrix $W \in \mathbb{R}^{m \times n}$, RTN computes per-group scales $s = \max |W| / (2^{b-1} - 1)$ and quantizes by $W_q = s \cdot \text{clamp}(\text{round}(W/s), -q_{\max}, q_{\max})$. The relative reconstruction error is $\|W - W_q\|_F / \|W\|_F$.

We chose RTN as the primary method because its reconstruction error depends only on the weight distribution, not on calibration data, giving the cleanest test of the recon-error/sensitivity relationship. We add GPTQ to test whether component-level patterns survive a calibration-aware quantizer.

\subsection{Sensitivity Measurement}

For each (model, method) pair we (1)~compute a baseline WikiText-2~\cite{merity2017pointer} perplexity over 16,384 tokens in 1,024-token blocks; (2)~for each attention projection independently, quantize it to $b \in \{3,4\}$ bits while keeping all other weights at full precision; (3)~record reconstruction error, the resulting perplexity, $\Delta\text{PPL} = \text{PPL}_q - \text{PPL}_{\text{base}}$, and a per-projection activation-weighted quantization error $\hat{S}=m^{-1}\sum_{i=1}^{m}\sum_{j=1}^{n}(W_{ij}-W_{q,ij})^2\widehat{\mathbb{E}}[x_j^2]$. Activation second moments are estimated from the WikiText-2 evaluation sequence for RTN and from the 2,048-token C4 calibration sequence for GPTQ. The full sweep yields a maximum of 256 measurements per pair.

\section{Results}

\subsection{Reconstruction Error Is a Weak Within-Component Predictor}
\label{sec:weak}

If reconstruction error were a reliable proxy for functional sensitivity, then within a single component type we would expect projections with higher reconstruction error to show larger $\Delta\text{PPL}$. Table~\ref{tab:within_r2} reports the within-component $R^2$ at 3-bit RTN.

\begin{table}[t]
\centering
\caption{Within-component $R^2$ between reconstruction error and $\Delta\text{PPL}$ at 3-bit RTN. Bold marks the only cell exceeding 0.50. 27 of 36 cells are below 0.10; overall median is 0.044.}
\label{tab:within_r2}
\small
\begin{tabular}{lcccc}
\toprule
\textbf{Model} & \textbf{Q} & \textbf{K} & \textbf{V} & \textbf{O} \\
\midrule
GPT-J-6B & 0.003 & 0.007 & 0.002 & 0.397 \\
LLaMA-1-7B & 0.027 & 0.092 & \textbf{0.560} & 0.013 \\
LLaMA-2-7B & 0.051 & 0.011 & 0.363 & 0.096 \\
LLaMA-3-8B & 0.114 & 0.006 & 0.064 & 0.027 \\
Mistral-7B & 0.000 & 0.038 & 0.056 & 0.000 \\
OPT-1.3B & 0.114 & 0.428 & 0.047 & 0.039 \\
OPT-6.7B & 0.040 & 0.009 & 0.128 & 0.137 \\
Qwen2.5-1.5B & 0.052 & 0.041 & 0.036 & 0.085 \\
Qwen2.5-7B & 0.001 & 0.155 & 0.024 & 0.047 \\
\midrule
\textit{Median} & 0.040 & 0.038 & 0.056 & 0.047 \\
\bottomrule
\end{tabular}
\end{table}

The pattern is stark. Of 36 cells, 27 (75\%) have $R^2 < 0.10$; only one cell (LLaMA-1 V projections) exceeds $R^2 = 0.50$. The overall median is $R^2 = 0.044$.

\subsection{The V-Projection Natural Experiment}

In several models the within-component test is unusually clean: reconstruction error is essentially constant across layers, so any variation in $\Delta\text{PPL}$ cannot in principle be predicted by it. Table~\ref{tab:cv_recon} reports the coefficient of variation of reconstruction error within each component.

\begin{table}[t]
\centering
\caption{Coefficient of variation (\%) of reconstruction error across layers within each component (3-bit RTN). Bold cells have CV $<$ 1.5\%.}
\label{tab:cv_recon}
\small
\begin{tabular}{lrrrr}
\toprule
\textbf{Model} & \textbf{Q} & \textbf{K} & \textbf{V} & \textbf{O} \\
\midrule
GPT-J-6B & \textbf{0.2} & \textbf{0.2} & \textbf{0.2} & \textbf{0.4} \\
LLaMA-1-7B & 4.7 & 4.6 & \textbf{0.5} & 1.9 \\
LLaMA-2-7B & 9.3 & 9.4 & \textbf{0.9} & 4.9 \\
LLaMA-3-8B & 3.0 & 4.9 & 3.4 & \textbf{1.3} \\
Mistral-7B & 2.4 & 3.0 & 2.9 & 1.7 \\
OPT-1.3B & 8.1 & 6.0 & \textbf{1.3} & 6.1 \\
OPT-6.7B & 4.3 & 3.8 & 1.8 & 5.5 \\
Qwen2.5-1.5B & 2.6 & 4.6 & 11.2 & \textbf{0.8} \\
Qwen2.5-7B & 2.1 & 3.2 & 8.8 & \textbf{0.8} \\
\bottomrule
\end{tabular}
\end{table}

In ten cells, CV of reconstruction error is below 1.5\%. Yet $\Delta\text{PPL}$ varies substantially across layers in every one of these cells---a clean falsification of the recon-as-sensitivity assumption.

\subsection{What Does Explain $\Delta\text{PPL}$?}
\label{sec:variance}

We compare three candidate explanatory variables: component type, categorical layer identity, and reconstruction error. Table~\ref{tab:variance} reports all three on 3-bit RTN data.

\begin{table}[t]
\centering
\caption{Marginal explained variance of $\Delta\text{PPL}$ at 3-bit RTN. Bold marks the strongest predictor per model.}
\label{tab:variance}
\small
\begin{tabular}{lccc}
\toprule
\textbf{Model} & $\boldsymbol{\eta^2_{\text{comp}}}$ & $\boldsymbol{\eta^2_{\text{layer}}}$ & $\boldsymbol{R^2_{\text{recon}}}$ \\
\midrule
GPT-J-6B & \textbf{0.340} & 0.150 & 0.008 \\
LLaMA-1-7B & 0.149 & \textbf{0.194} & 0.015 \\
LLaMA-2-7B & 0.160 & \textbf{0.285} & 0.001 \\
LLaMA-3-8B & 0.198 & \textbf{0.251} & 0.022 \\
Mistral-7B & 0.177 & \textbf{0.242} & 0.016 \\
OPT-1.3B & 0.079 & \textbf{0.358} & 0.012 \\
OPT-6.7B & 0.013 & \textbf{0.222} & 0.001 \\
Qwen2.5-1.5B & 0.202 & \textbf{0.204} & 0.017 \\
Qwen2.5-7B & \textbf{0.318} & 0.301 & 0.001 \\
\bottomrule
\end{tabular}
\end{table}

Both categorical signals beat reconstruction error in all 9 models. Layer identity is the strongest individual predictor in 7 of 9 models; component type is strongest in GPT-J-6B and Qwen2.5-7B. The picture: ``component type and layer identity both matter; reconstruction error mostly does not.''

\subsection{Architecture-Dependent Component Dominance}
\label{sec:dominance}

\begin{table}[t]
\centering
\caption{Percentage of total positive $\Delta\text{PPL}$ by component (3-bit RTN). Bold indicates the dominant component.}
\label{tab:dominance}
\small
\begin{tabular}{lccccl}
\toprule
\textbf{Model} & \textbf{Q\%} & \textbf{K\%} & \textbf{V\%} & \textbf{O\%} & \textbf{Dom.} \\
\midrule
GPT-J-6B & 14 & 19 & \textbf{48} & 19 & V \\
LLaMA-1-7B & 12 & 12 & \textbf{51} & 25 & V \\
LLaMA-2-7B & 13 & 12 & \textbf{50} & 25 & V \\
LLaMA-3-8B & 14 & 8 & \textbf{42} & 36 & V \\
Mistral-7B & 13 & 16 & \textbf{41} & 29 & V \\
OPT-1.3B & 12 & \textbf{39} & 34 & 14 & K \\
OPT-6.7B & 28 & 16 & \textbf{44} & 12 & V \\
Qwen2.5-1.5B & 12 & 15 & 32 & \textbf{42} & O \\
Qwen2.5-7B & 12 & 13 & \textbf{38} & 37 & V \\
\bottomrule
\end{tabular}
\end{table}

V projections are dominant in seven of nine models, including all three LLaMA variants. Q is not dominant in any model and has the smallest positive-sensitivity share in five of nine models.

\subsection{Does the Dominance Picture Survive GPTQ?}
\label{sec:gptq}

\begin{table}[t]
\centering
\caption{Dominant component under RTN vs.\ GPTQ at 3-bit. Match rate = 5/7.}
\label{tab:rtn_gptq}
\small
\begin{tabular}{lccc}
\toprule
\textbf{Model} & \textbf{RTN dom.} & \textbf{GPTQ dom.} & \textbf{Match} \\
\midrule
GPT-J-6B & V (48\%) & V (33\%) & $\checkmark$ \\
LLaMA-2-7B & V (50\%) & O (39\%) & --- \\
LLaMA-3-8B & V (42\%) & V (43\%) & $\checkmark$ \\
Mistral-7B & V (41\%) & V (46\%) & $\checkmark$ \\
OPT-1.3B & K (39\%) & K (38\%) & $\checkmark$ \\
OPT-6.7B & V (44\%) & V (32\%) & $\checkmark$ \\
Qwen2.5-1.5B & O (42\%) & V (36\%) & --- \\
\midrule
Match rate & & & 5/7 \\
\bottomrule
\end{tabular}
\end{table}

Five of seven models retain the same dominant component under GPTQ. The two mismatches are shifts within \{K, V, O\} rather than to Q.

\subsection{Activation-Weighted Error Is a Better V-Projection Predictor}
\label{sec:hessian}

\begin{table}[t]
\centering
\caption{Within-component $R^2$ between $\log \hat{S}$ and $\Delta\text{PPL}$ at 3-bit RTN. Bold marks cells with $R^2 \ge 0.40$. V column median $R^2 = 0.20$ vs.\ 0.06 for reconstruction error.}
\label{tab:hsens_r2}
\small
\begin{tabular}{lcccc}
\toprule
\textbf{Model} & \textbf{Q} & \textbf{K} & \textbf{V} & \textbf{O} \\
\midrule
GPT-J-6B & 0.002 & 0.006 & 0.202 & 0.198 \\
LLaMA-1-7B & 0.045 & 0.048 & \textbf{0.439} & 0.048 \\
LLaMA-2-7B & 0.097 & 0.012 & \textbf{0.490} & 0.067 \\
LLaMA-3-8B & 0.017 & 0.001 & \textbf{0.632} & 0.002 \\
Mistral-7B & 0.018 & 0.063 & \textbf{0.514} & 0.023 \\
OPT-1.3B & 0.010 & 0.037 & 0.028 & 0.280 \\
OPT-6.7B & 0.043 & 0.007 & 0.088 & 0.003 \\
Qwen2.5-1.5B & 0.073 & 0.041 & 0.168 & 0.341 \\
Qwen2.5-7B & 0.161 & 0.129 & 0.038 & 0.009 \\
\midrule
\textit{Median} & 0.043 & 0.037 & \textbf{0.202} & 0.048 \\
\bottomrule
\end{tabular}
\end{table}

Activation-weighted quantization error is a meaningfully better V-projection predictor: median $R^2$ across V cells is 0.20 (vs.\ 0.06 for reconstruction error), and four V cells exceed $R^2 = 0.40$. This is consistent with the picture that V's contribution to attention output is roughly linear ($\text{Attn}(Q,K,V) = \text{softmax}(QK^\top/\sqrt{d}) V$), so weighting quantization error by observed input second moments can track $\Delta\text{PPL}$ more accurately for V than for K or Q, where the softmax nonlinearity intervenes.

\section{Discussion}

\textbf{What the data does and does not say.} The strong negative finding is robust: reconstruction error is a weak within-component predictor of $\Delta\text{PPL}$ in 27 of 36 cells. Two stronger signals---component type and layer identity---carry more information, but neither is dominant across all models. The most consistent pattern is V dominance in seven of nine models.

\textbf{Implications for sensitivity-aware mixed precision.} First, allocation policies that rely on \emph{within-component} reconstruction-error rankings will be poorly supported by reconstruction error in 75\% of the model--component cases. Second, an allocation policy that simply protects V projections (e.g., 4 bits for V, 3 bits for everything else) is a plausible default to evaluate for LLaMA-family models.

\textbf{From V dominance to practical allocation rules.} The seven-of-nine V dominance result suggests a simple, architecture-indexed heuristic for mixed-precision attention quantization: assign V projections one additional bit relative to Q, K, and O. For a model with 32 layers and four projections per layer, this affects only 25\% of attention weight matrices while targeting the component that contributes 38--51\% of total positive $\Delta\text{PPL}$ in the V-dominant models. The cost is modest: in a 7B-parameter model where attention projections account for roughly one-third of total parameters, upgrading V from 3 to 4 bits increases the average attention bit-width from 3.0 to 3.25 and the overall model bit-width by approximately 0.08 bits. Whether this marginal storage cost is justified depends on the magnitude of the perplexity recovery, which we leave to future work to measure in a full mixed-precision deployment. Importantly, our data also show that this heuristic is not universal: OPT-1.3B is K-dominant and Qwen2.5-1.5B is O-dominant, so any production system should include a lightweight profiling step---quantizing one projection per component at a single bit-width on a small calibration set---to identify the dominant component before committing to an allocation policy.

\textbf{Relationship between attention type and dominance.} The nine models in our study span two attention mechanisms: multi-head attention (MHA) and grouped-query attention (GQA). One might expect GQA models to exhibit different dominance patterns because their K and V projections are smaller (fewer heads), changing both the parameter count and the per-head redundancy available to absorb quantization error. Our data do not support a clean separation along this axis: V is dominant in MHA models (GPT-J, LLaMA-1, LLaMA-2, and OPT-6.7B) and GQA models (LLaMA-3, Mistral, Qwen2.5-7B). The two non-V-dominant models are split across MHA (OPT-1.3B, K-dominant) and GQA (Qwen2.5-1.5B, O-dominant). This suggests that attention type alone does not determine the dominant component, and that other architectural choices---such as the presence or absence of bias terms, the use of rotary versus learned positional embeddings, or the ratio of attention to FFN parameters---may interact with quantization sensitivity in ways that merit further investigation.

\textbf{Why V?} We hypothesize two contributing factors. First, V acts roughly linearly in the attention output: errors in V propagate without the dampening effect of the softmax that K and Q errors are subject to. Second, V projections tend to have unusually narrow weight distributions, yielding flat per-layer reconstruction-error profiles and meaning that any error is structurally hard to compensate for.

\textbf{Limitations.} Our evaluation is restricted to RTN and GPTQ. WikiText-2 perplexity over 16,384 tokens is a coarse functional metric---replicating with downstream task accuracy (HellaSwag~\cite{zellers2019hellaswag}, ARC~\cite{clark2018arc}, PIQA~\cite{bisk2020piqa}) would strengthen the conclusions. We use diagonal activation-second-moment weighting rather than the full loss Hessian; richer second-order information might improve V predictions further.

\section{Conclusion}

We have shown, on 3,808 single-projection measurements across nine RTN models and seven GPTQ models, that the relative weight reconstruction-error signal examined here is a weak within-component predictor of perplexity sensitivity (median $R^2 = 0.044$). Component type and layer identity both carry more information. Among components, V projections dominate in seven of nine models, including all three LLaMA variants, and the dominance pattern is largely preserved between RTN and GPTQ. Activation-weighted quantization error is a meaningfully better V-projection predictor than reconstruction error (median $R^2$ of 0.20 vs.\ 0.06), suggesting that practical sensitivity-aware mixed-precision schemes should privilege V projections specifically and use activation-weighted scoring rather than reconstruction-error-based scoring within components.

\section*{Reproducibility}

All experiments use publicly available models from Hugging Face. Experiments were conducted on NVIDIA H100 PCIe GPUs (80 GB), CUDA 12.1. Perplexity is evaluated on 16,384 tokens from WikiText-2 in non-overlapping 1,024-token blocks. GPTQ calibration uses 2,048 tokens from C4, while RTN activation-weighted scores use the WikiText-2 evaluation activations; both methods use group size 128. GPTQ and RTN are deterministic given fixed weights and token sequences. Code to reproduce all experiments and analyses is publicly available at \url{https://github.com/Kasun-Dewage/What-predicts-Quantization-Sensitivity-2026}.

\bibliographystyle{IEEEtran}

\end{document}